\documentclass[10pt,twocolumn,letterpaper]{article}

\usepackage{cvpr}              

\usepackage{graphicx}
\usepackage{amsmath}
\usepackage{amssymb}
\usepackage{booktabs}
\usepackage[accsupp]{axessibility}  

\usepackage[pagebackref,breaklinks,colorlinks]{hyperref}

\usepackage[capitalize]{cleveref}
\crefname{section}{Sec.}{Secs.}
\Crefname{section}{Section}{Sections}
\Crefname{table}{Table}{Tables}
\crefname{table}{Tab.}{Tabs.}

\def\cvprPaperID{14} 
\def\confName{CVPR}
\def\confYear{2022}

\begin{document}

\title{Deep Image Retrieval is not Robust to Label Noise}

\author{Stanislav Dereka\\
Tinkoff\\
{\tt\small st.dereka@gmail.com}
\and
Ivan Karpukhin\\
Tinkoff\\
{\tt\small i.a.karpukhin@tinkoff.ru}
\and
Sergey Kolesnikov\\
Tinkoff\\
{\tt\small scitator@gmail.com}
}
\maketitle

\begin{abstract}
    Large-scale datasets are essential for the success of deep learning in image retrieval. However, manual assessment errors and semi-supervised annotation techniques can lead to label noise even in popular datasets. As previous works primarily studied annotation quality in image classification tasks, it is still unclear how label noise affects deep learning approaches to image retrieval. In this work, we show that image retrieval methods are less robust to label noise than image classification ones. Furthermore, we, for the first time, investigate different types of label noise specific to image retrieval tasks and study their effect on model performance.
\end{abstract}


\section{Introduction}
Over the last decade, deep learning achieved impressive results in multiple computer vision domains, including image classification \cite{krizhevsky2012imagenet}, image retrieval \cite{babenko2014cnnretrieval}, and face verification \cite{deng2019arcface}.
One reason behind the success of deep learning is the ability of deep neural networks to extract useful information from large amounts of annotated data.
It was shown that increasing the amount of data along with model size leads to an improvement in prediction quality \cite{joulin2016weaklysupervised,sun2017unreasonable}. To do so, datasets with tens of thousands and even millions of items were collected for deep learning \cite{guo2016msceleb,kuznetsova2020openimages}. While the majority of data can usually be obtained from the Internet, annotating this data is the most laborious part of the data preparation process \cite{ILSVRC15}.
When provided with millions of items, it is almost impossible to manually annotate all of them.
Multiple semi-supervised approaches were created to reduce the amount of manual work required in data annotation \cite{nech2017level}. However, both manual annotation and semi-supervised algorithms can introduce errors to final results. It was shown recently that the annotation error rate in large-scale datasets can exceed 40\% \cite{wang2018devil}.

Previous research on label noise in computer vision was primarily focused on image classification \cite{patrini2017asymmetric, algan2021labelnoisesurvey}.
%
%
In this paper, we study the performance of deep convolutional neural networks in image classification and retrieval tasks when working with label noise. The core contributions of the paper can be summarized as follows:
\begin{enumerate}
    \item We study the effects of label noise in image retrieval on Stanford Online Products \cite{oh2016sop}, In-shop \cite{liu2016inshop} and face recognition datasets \cite{guo2016msceleb, huang2014lfw} along with image classification on ImageNet \cite{ILSVRC15}.
    Our results show that image retrieval models are less robust to label noise than image classification models under similar training conditions.
    \item We study different retrieval-specific noise types with respect to the number of mislabeled samples and corrupted classes. Our findings show that the amount of corrupted classes affects performance more than the proportion of corrupted samples in each class. 
\end{enumerate}

\section{Related Work}

In deep computer vision, label noise was primarily studied in the context of image classification \cite{patrini2017asymmetric, algan2021labelnoisesurvey}. Many techniques were proposed to reduce the effects of label noise on the final models' performance. For example, one of these methods involves automatically detecting mislabeled items and excluding them from training \cite{pleiss2020cleanremove, sharma2020noiserank,kuang2022efficient}.
Another way to deal with label noise is to design specialized training algorithms \cite{castells2020superloss,liu2021towards,damodaran2020entropic}. While these methods can, to some degree, reduce the destructive impact of annotation errors, they are still affected by the labels' quality. While we find these approaches valuable for the community, we argue for deeper model robustness analysis in image retrieval. 

Several previous works addressed the general question of deep image classification robustness to label noise \cite{joulin2016weaklysupervised,krause2016unreasonable,sun2017unreasonable,yi2019noisecorrection}. It was shown that modern convolutional neural network (CNN) architectures, such as ResNet \cite{he2016resnet}, are robust to label noise in image classification. While the models' final accuracy reduces with the growing noise level, the model still trains even with five times more noisy samples than clean ones. Similar studies were recently performed for image retrieval and face recognition \cite{wang2018devil,liu2021mlnoise,ibrahimi2022irnoise,corbiere2017leveraging}, showing that metric learning algorithms are more sensitive to label noise than classification methods. While there is a fundamental difference between classification and metric learning approaches when it comes to the effects of label noise, the reason for this difference is still unclear. To address the issue of earlier works using different network architectures, training strategies, and noise types, we compared retrieval, classification, and face recognition methods under similar training conditions. Also, we study novel types of noising schemes designed for image retrieval.

\section{Label Noise in Image Retrieval}\label{sec:noise}



Many image retrieval datasets are gathered using unreliable methods such as search engine crawling \cite{corbiere2017leveraging, wang2018devil, guo2016ms}, which leads to label noise and degrades learning \cite{wang2018devil}. Manual cleaning of large-scale datasets with millions of samples can be expensive and resource demanding \cite{wang2018devil}. According to previous research, data annotation techniques used in image retrieval can produce specific types of label noise \cite{wang2018devil,liu2021mlnoise}. In particular, noise can be concentrated in clusters of similar images leading to non-uniform distribution of noisy labels among classes. Moreover, well-studied datasets, e.g. Labeled Faces in the Wild (LFW) \footnote{See known LFW errors on the official website: \url{http://vis-www.cs.umass.edu/lfw/}} \cite{huang2014lfw}, include "trash" classes which completely consist of mislabeled samples. We hereby raise a question of how the number of corrupted (trash) classes affects final model performance? We call class corrupted if almost all items in this class have the wrong label. This question differs from previous research in image classification, where some amount of clean data needs to be provided for each class.

\begin{figure}[h]
  \centering
  \includegraphics[width=\linewidth]{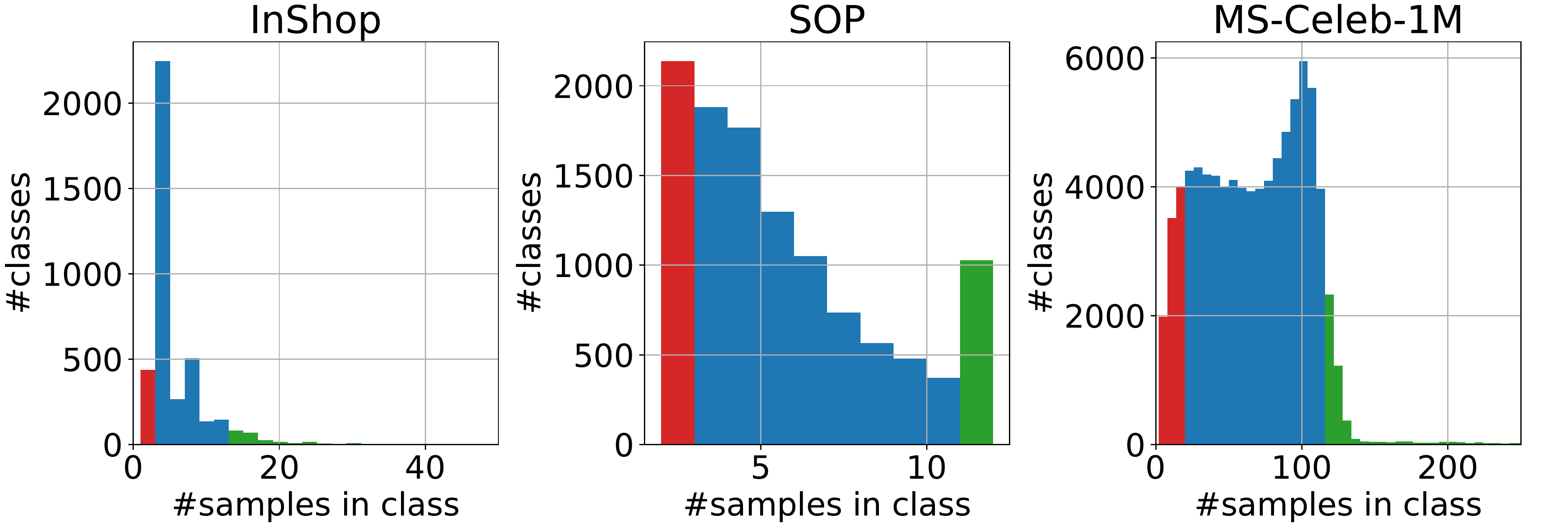}
  \caption{Class size distributions in popular image retrieval datasets. Examples of classes affected by small and large class label noise are highlighted in red and green respectively.}
  \label{fig:imbalance}
\end{figure}

When modeling label noise patterns, there are multiple ways to choose classes for corruption. As shown in Figure \ref{fig:imbalance}, image retrieval datasets are highly class-imbalanced. Large classes can contain thousands of times more samples than small ones. By selecting a fixed amount of noisy items, we can choose either many small classes or a small number of large classes. Based on this observation, we propose two new label noising schemes for image retrieval.

In the proposed \textbf{large class label noise} we select the most frequent classes for a given proportion of samples in a dataset and shuffle the labels among the samples of the selected classes. The number of classes to corrupt is selected to match the expected number of corrupted samples. \textbf{Small class label noise} is similar to large class label noise, but the rarest classes are corrupted instead of large ones. The dataset parts which were corrupted by each noise pattern are illustrated on histograms in Figure \ref{fig:imbalance}.
The proposed patterns are opposites regarding the number of totally corrupted classes. Large class label noise minimizes the number of corrupted classes while the number of corrupted samples is fixed. In contrast, small class label noise maximizes the number of corrupted classes. 


As a baseline, we study traditional \textbf{uniform label noise} \cite{algan2021labelnoisesurvey}, also known as label flipping \cite{wang2018devil}. 
In this noise pattern, a given proportion of samples in a dataset is randomly selected, and a random label from all the available labels is assigned to the selected samples. This noise pattern slightly corrupts each class in the dataset with a probability proportional to the number of samples in a class. 

\section{Experiment Setup}
\label{sec:setup}

\subsection{Datasets and Evaluation Metrics}



The scope of this paper covers both image retrieval and image classification datasets. We include the image classification task in our work to compare its robustness to label noise with image retrieval. Closed-set classification datasets share the same set of labels between training and testing. The goal of an algorithm in this case is to predict label for each input image. Accuracy is usually used to evaluate classification performance.
For \textit{image classification} we use the ImageNet \cite{ILSVRC15}.

The goal of \textit{image retrieval (IR)} is to find images from the gallery collection which are similar to the query image \cite{oh2016sop, musgrave2020metric}. Let's call retrieval successful for some query if the most similar image in the gallery has the same label. We measure Recall@1 (R@1) during evaluation, which is equal to the fraction of successful retrievals.

Datasets for image retrieval usually have class-disjoint training and testing parts. This setup is also called open-set problem. In our experiments we use Stanford online products (SOP) \cite{song2016deep} and In-shop Clothes (InShop) \cite{liu2016inshop} datasets. Both of these datasets have severe class imbalance as shown in Figure \ref{fig:imbalance}.

In addition to the above, we extend the scope of our work by adding \textit{face recognition (FR)} datasets. 
The majority of face recognition datasets has class imbalance similar to other retrieval datasets. Furthermore, face recognition tasks are open-set and can be subject to common image retrieval problems \cite{WANG2021215}. Finally, public FR datasets are large-scale and make it possible for the experiment setup to be close to real-world practical applications. For training, we use the MS-Celeb-1M \cite{guo2016msceleb} dataset, which contains over 10M images of around 100K individuals. For face recognition evaluation, we chose the LFW \cite{huang2014lfw} dataset. In addition to computing R@1, we evaluate \textit{face verification} performance. LFW verification test set includes 6K image pairs labeled with 0 (same person) or 1 (different people). During testing, a similarity in model embedding space is computed for each pair. For face verification, we then compute TPR@FPR metric for 10$^{-3}$ FPR by using a common FR approach \cite{WANG2021215, kemelmacher2016megaface}.

\subsection{Training Details}



In image retrieval and classification datasets (SOP, In-shop, ImageNet) we preprocess images using approaches from previous papers \cite{he2016resnet, musgrave2020metric}. Images in ImageNet, SOP, and InShop datasets are resized to 256 pixels by the shortest side. During training, each image is randomly cropped to 224x224 pixels. During testing, we use central crop. In MS-Celeb-1M v2 dataset \cite{deng2019arcface}, the size of aligned images is 112x112 pixels in both training and testing. We use random horizontal flip augmentation for all datasets.

In all experiments, we use ResNet50 \cite{he2016resnet} CNN architecture as a backbone network. ResNet50 is initialized with ImageNet pretrain weights for all the tasks, except for ImageNet and MS-Celeb-1M, where random initialization is used. The embedding size is set to 512 in all experiments.

Our models are trained with the ArcFace \cite{deng2019arcface} loss function. According to recent works \cite{musgrave2020metric, deng2019arcface}, this is one of the best-performing loss functions in considered benchmarks. We use the default ArcFace scale and margin hyperparameters from the original paper for training on SOP, InShop, and MS-Celeb-1M. For ImageNet, we use Normalized Softmax loss \cite{wang2017normface}.

We use the stochastic gradient descent (SGD) optimizer with momentum 0.9 and weight decay $0.0001$. The initial learning rate is set to 0.1. We follow the pipeline from \cite{deng2019arcface} and perform 16 epochs of training, decreasing the learning rate by a factor of 0.1 at the 9th and 14th epochs. In all experiments we set batch size to 256.

\section{Experiment Results}

\subsection{Robustness Comparison of Classification and Retrieval Tasks}

\begin{table}[h]
\begin{tabular}{@{}cccccc@{}}
\toprule
Dataset & Metric & \multicolumn{4}{c}{Label noise level} \\
& & clean & 0.01   & 0.05  & 0.1  \\
\midrule
ImageNet & Accuracy & 67.32 & 67.09 & 66.42 & 65.80 \\
InShop & R@1 & 84.61 & 82.09 & 74.70 & 72.70 \\
SOP & R@1 & 63.49 & 63.68 & 60.72 & 58.78 \\
LFW & R@1 & 67.22 & 67.36 & $\approx$0.0 & $\approx$0.0 \\
LFW & TPR@10$^{-3}$ & 99.27 & 99.23 & $\approx$0.0 & $\approx$0.0 \\
\bottomrule
\end{tabular}
\caption{Uniform label noise effects on model performance for image classification (ImageNet) and verification tasks compared to image retrieval tasks. All metric values are shown in \%. We label cases where the model was unable to generalize on test set with $\approx$.}
\label{tab:cls_vs_ir}
\end{table}


The goal of our experiments is to measure the robustness of image retrieval tasks for comparison with image classification. We quantify robustness as a drop in performance when a model is trained on a noisy dataset compared to training on a clean one. 
We compare models trained with multiple noise levels ranging from 0 (clean dataset) to 10\%.
The results are presented in Table \ref{tab:cls_vs_ir}.

\begin{figure}[h]
  \centering
  \includegraphics[width=\linewidth]{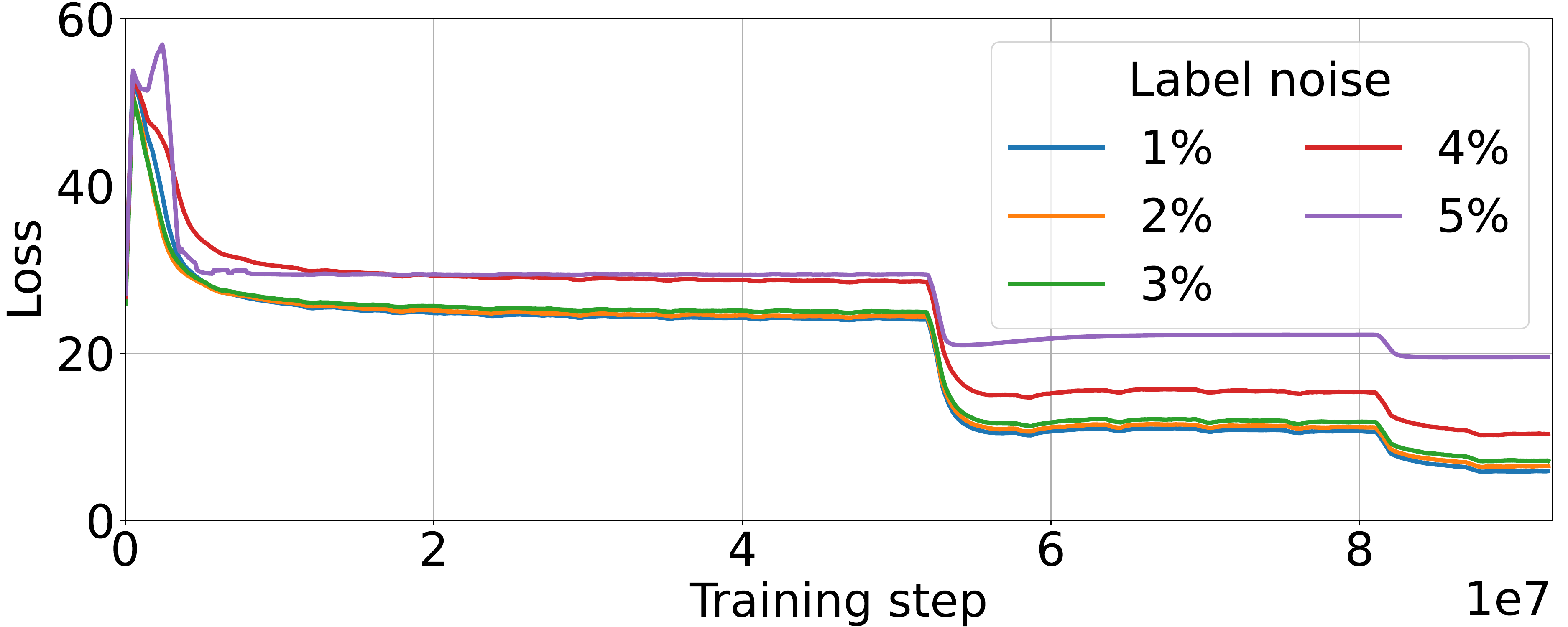}
  \caption{Training curves for MS-celeb-1M dataset with different label uniform noise levels.}
  \label{fig:train_curves}
\end{figure}

On all retrieval tasks, we observe a rapid decrease of quality with the increase of uniform label noise level. Even 5\% of label noise causes a relative performance drop of up to 10\%. In contrast, classification accuracy under equal conditions shows a much smaller drop. These results show that image classification is more robust to uniform label noise than image retrieval.

One of the most interesting findings from our experiments is that, when a model is trained on the MS-Celeb-1M dataset with at least 5\% label noise, its score is close to that of an untrained model. In order to clarify this phenomena, we show training curves for several noise levels in Figure \ref{fig:train_curves}. It can be seen that at 5\% label noise, loss function optimization still occurs, which means that the model training still converges. These findings allow us to conclude that image retrieval setups are extremely sensitive to label noise and their performance can be affected drastically by only a small fraction of label noise.

\subsection{Effects of Noise Patterns}


\begin{table}[h]
    \begin{tabular}{ccccc}
    \toprule
    Noise type & Dataset & \multicolumn{3}{c}{Noise level} \\
    & & 0.01 & 0.05 & 0.1 \\
    & & \multicolumn{3}{c}{Corrupted classes, \%} \\
    \midrule
    Uniform & InShop & 6.0 & 26.0 & 44.3 \\
    Large Classes & InShop & 0.05 & 0.5 & 1.6 \\
    Small Classes & InShop & 2.5 & 11.1 & 19.2 \\
    Uniform & SOP & 5.0 & 22.6 & 40.0 \\
    Large Classes & SOP & 0.4 & 2.2 & 4.4 \\
    Small Classes & SOP & 2.6 & 13.2 & 23.8 \\
    Uniform & MS-Celeb-1M & 45.5 & 88.0 & 95.5 \\
    Large Classes & MS-Celeb-1M & 0.2 & 1.4 & 4.2 \\
    Small Classes & MS-Celeb-1M & 7.3 & 19.5 & 29.1 \\
    \bottomrule
    \end{tabular}
    \caption{Percentage of corrupted classes for each dataset and noise pattern for three noise levels. A class is considered corrupted if at least one label in this class has been flipped during the noising procedure.}
    \label{tab:corrupted_ids}
\end{table}

According to the previous experiment, image retrieval is not robust to label noise. Therefore, we can raise a question of how the number of corrupted classes affects retrieval performance. To answer this question, we corrupted a training set using new types of label noise for image retrieval, previously described in section \ref{sec:noise}. Given a fixed set of noisy items, we can either corrupt a small number of large classes or many small classes. These noising approaches are called large-class noise and small-class noise respectively. The total amount of corrupted items and corrupted classes in our experiments is given in Table \ref{tab:corrupted_ids}.

\begin{figure}[h]
  \centering
  \includegraphics[width=\linewidth]{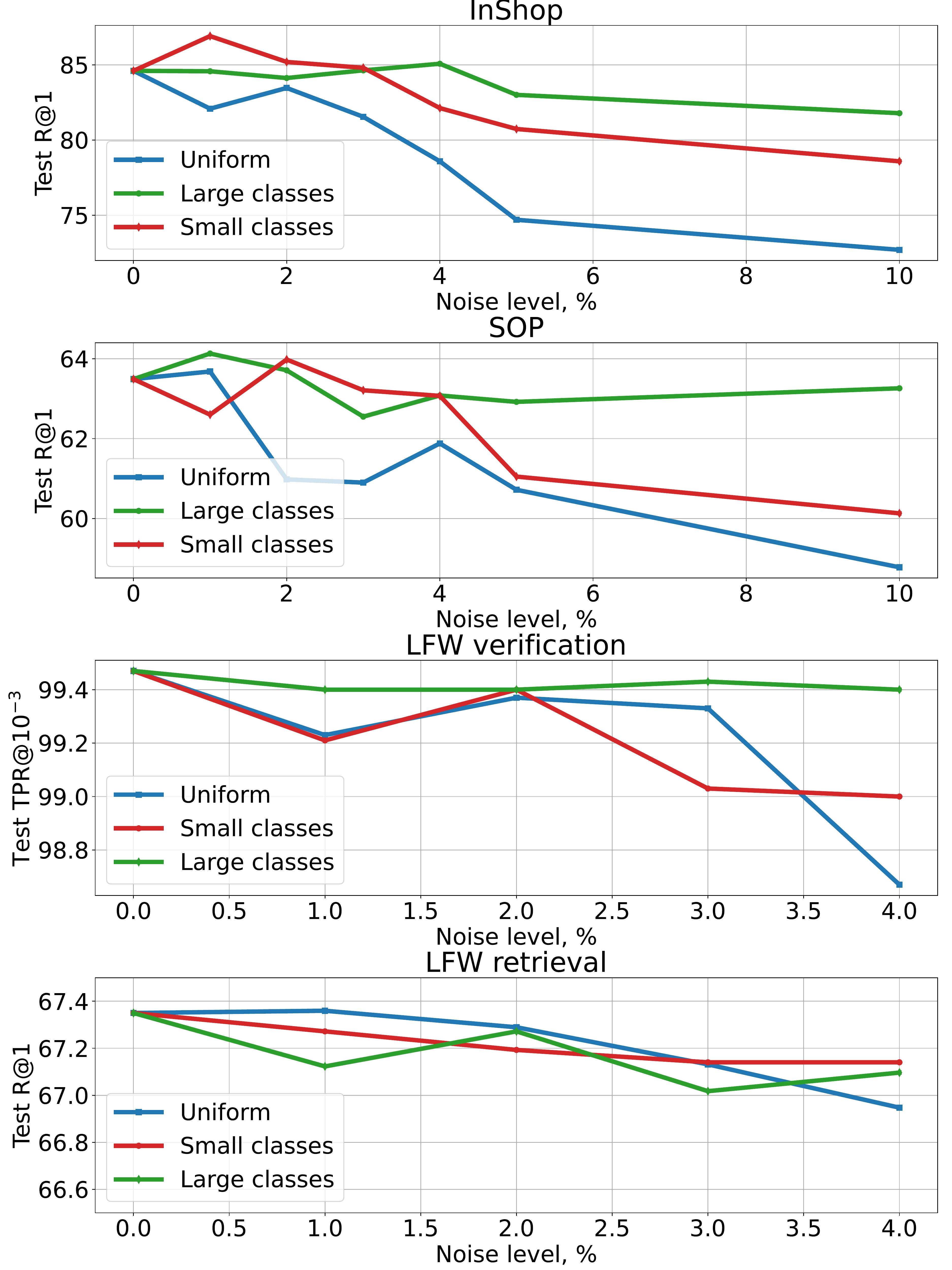}
  \caption{Label noise patterns' effect on test performance.}
  \label{fig:metrics}
\end{figure}

It can be seen from Figure \ref{fig:metrics} that the more training set classes are affected by label noise, the higher is the drop in performance, even if the total amount of corrupted elements doesn't change. For example, the noise level 5\% on the InShop dataset leads to a 10\% quality drop for uniform noise with 26\% corrupted classes. On the other hand, large and small class noises with the same 5\% noise level produce only 2\% and 4.5\% quality drop, while the numbers of corrupted classes for these noises are 0.05\% and 2.5\% respectively (Table \ref{tab:corrupted_ids}).

As a conclusion of these experiments, small class label noise leads to more significant degradation than large class label noise. The effect of uniform label noise is the strongest, as it affects more classes than the other noise types. It's worth noting that thoroughly corrupted classes affect performance less than a large amount of slightly damaged classes.

\section{Discussion and Future Work}

According to our experiments, image retrieval is less robust to label noise than image classification. We used similar models, training objectives, and optimization methods for both tasks. The difference in robustness can lie in contrast between closed-set and open-set tasks. Image retrieval models need to generalize well to unseen classes, while classification tasks share the same set of labels between training and testing. Our experiments show that while models converge with all considered noise levels, the test set accuracy completely drops even for 5\% label noise. 
These results bring us to the conclusion that image retrieval models fail to generalize to unseen classes in this case, highlighting the need for further investigation on label noise robustness in image retrieval and other open-set tasks.




The goal of this work is to highlight difference between classification and retrieval robustness to label noise as well as to study retrieval-specific types of annotation errors. We choose ArcFace as one of the best-performing models for our experiments. Future work can extend presented results to a variety of training objectives \cite{weinberger2009triplet, movshovitz2017proxynca} and model architectures \cite{szegedy2017inception, dosovitskiy2020vit}.


\section{Conclusion}
In this work, we studied image retrieval robustness to label noise. According to our experiments, image retrieval is less robust to label noise than image classification approaches. Therefore, more attention must be paid to training data quality for image retrieval tasks. We further showed that the performance is rather affected by the number of corrupted classes than by the proportion of corrupted samples in each class.

{\small
\bibliographystyle{ieee_fullname}
\bibliography{main}
}

\end{document}